\documentclass[final]{cvpr}

\usepackage{times}
\usepackage{epsfig}
\usepackage{graphicx}
\usepackage{amsmath}
\usepackage{amssymb}
\usepackage{caption}
\usepackage{subcaption}
\usepackage{xcolor}
\usepackage[accsupp]{axessibility}

\usepackage{comment}
\usepackage{float}

\usepackage[pagebackref=true,breaklinks=true,colorlinks,bookmarks=false]{hyperref}

\def\DatasetName{RDE\xspace}

\def\cvprPaperID{37} 
\def\confYear{CVPR 2021}

\begin{document}

\title{ 
 Investigating Neural Architectures by Synthetic Dataset Design 
}

\author{Adrien Courtois\thanks{This work was supported by grants from Région Ile-de-France.}, Jean-Michel Morel, Pablo Arias\\
Centre Borelli, ENS Paris-Saclay\\
4 Av. des Sciences, 91190 Gif-sur-Yvette, France\\
{\tt\small adrien.courtois@ens-paris-saclay.fr}
}

\maketitle

\begin{abstract}

  Recent years have seen the emergence of many new neural network structures (architectures and layers).  To solve a given task, a network requires a certain set of abilities reflected in its structure. The required abilities depend on  each task.
 There is so far no systematic study of the real capacities of the proposed neural structures. The question of what each structure can and cannot achieve is only partially answered by its performance on common benchmarks. Indeed, natural data contain complex unknown statistical cues. It is therefore impossible to know what cues a given neural structure is taking advantage of in such data.
  In this work, we sketch a methodology to measure the effect of each structure on a network's ability, by designing \textit{ad hoc} synthetic datasets.
  Each dataset is tailored to assess a given ability and is reduced to its simplest form: each input contains exactly the amount of information needed to solve the task.
  We illustrate our methodology by building three datasets to evaluate each of the three following network properties: a) the ability to link local cues to distant inferences, b) the translation covariance and c) the ability to group pixels with the same characteristics and share information among them. Using a first simplified depth estimation dataset, we pinpoint a serious nonlocal deficit of the U-Net. We then evaluate how to resolve this limitation by embedding its structure with nonlocal layers, which allow computing complex features with long-range dependencies. Using a second dataset, we compare different positional encoding methods and use the results to further improve the U-Net on the depth estimation task. The third introduced dataset serves to demonstrate the need for self-attention-like mechanisms for resolving  more realistic depth estimation tasks.

\end{abstract}

\section{Introduction}


Deep learning has been characterized by significant advances in  fields ranging from computer vision \cite{paul2015review} to protein structure prediction \cite{jumper2021highly}. However, neural networks lack interpretability, and it is nearly impossible to predict the performance of a given structure on a task. While most of the effort is directed towards the explainability of the models themselves, the possibility that a better understanding of deep learning methods could come from better designed datasets has received little attention. In this work, we investigate this hypothesis by introducing a methodology to enhance the impact of architectural choices and to identify their flaws. 

Datasets of natural images need to be huge in order to capture the semantic complexity of the real world. While such datasets are necessary to ensure generalization to real world applications, their structure and information content is fully out of control.  The information given to the network can be ambiguous, sometimes contradictory, and the spatial interaction of features can be guided by hidden statistical dependences.  It is therefore hard or impossible to anticipate or explain the success or failure of a given network structure. 
A second ambiguity resides in the fact that each datum might not contain  enough information to solve the prescribed task. A third  ambiguity resides in the input itself: a plethora of semantic local and nonlocal cues coexist within the same image, which makes it difficult for an external observer to pinpoint the cause of success or failure of a given network structure.

A better understanding of neural networks requires characterizing their capabilities and linking them to their structure. To this end, we propose to train neural networks on datasets where those ambiguities have been lifted. That way, the success of a structure on a given task can only be attributed to the structure having a certain property, and not to some other uncontrolled statistics.
Alleviating the three sorts of ambiguities requires resorting to synthetic datasets. In this work, we introduce a methodology to design such unambiguous synthetic datasets to explore the properties of neural networks.

We illustrate this methodology on three datasets. First, we design a depth estimation task - the Rectangle Depth Estimation (RDE) dataset - to assess the non-local properties of the U-Net which, according to several authors \cite{qin2020u2, zhu2020map}, seems to be unable to exploit its large receptive field. In particular, we find that endowing the U-Net with nonlocal layers helps improve its nonlocal capability,  especially when a variant of the Lambda layer \cite{bello2021lambdanetworks} is used. Then, observations of the failure cases of the resulting structure raise the question of the positional encoding used within the Lambda layer. This leads us to design a second dataset, which aims at assessing the properties of the positional encoding. This second task allows us to design a better positional encoding method, which we successfully transfer to the first task. Finally, we design a dataset to evaluate the ability to group pixels with the same characteristics and share information among them. In particular, we find that self-attention \cite{vaswani2017attention} excels at this task.

The contributions of this paper are as follows:

\smallskip\noindent\emph{1)} We introduce a methodology to design synthetic datasets to be used to evaluate networks' properties. This allows to investigate neural architectures to better understand their capabilities. This methodology can be applied on any structure and for any data modality. 

\smallskip\noindent\emph{2)} We apply this methodology to evaluate three different network properties, namely: the ability to link local cues to distant inferences, the translation covariance and the ability to group pixels with the same characteristics and share information among them. For each property, a dataset is designed. These datasets can be used to evaluate any structure.

\smallskip\noindent\emph{3)} The datasets are used to compare and discuss multiple structures. The first dataset allows us to find a nonlocal deficit in the U-Net and to partially fix it by adding nonlocal layers in its structure. Then, the second dataset helps us find a way to incorporate positional encoding in the Lambda layer while ensuring translation equivariance. Finally, experiments on the third datasets point out that self-attention and variants excel at grouping pixels with the same characteristics and share information among them. The conclusions we draw on structures might lead to some improvement when handling real datasets, but this is not the goal of this paper. Rather, the proposed methodology may be used to verify unambiguously the effect of each  proposed neural structure on \textit{ad hoc}  synthetic data.

\begin{figure}[tb!]
    \centering
    \begin{subfigure}[b]{0.38\linewidth}
        \centering
        \includegraphics[width=\linewidth]{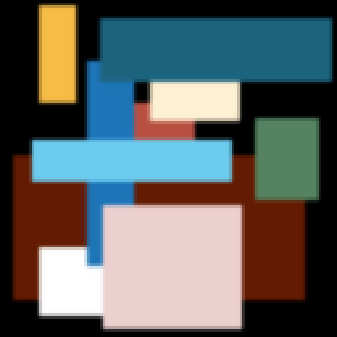}
    \end{subfigure}
    \begin{subfigure}[b]{0.38\linewidth}
        \centering
        \includegraphics[width=\linewidth]{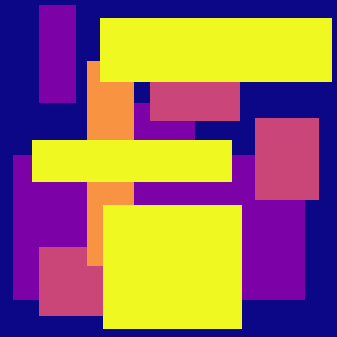}
    \end{subfigure}
    \begin{subfigure}[b]{0.38\linewidth}
        \centering
        \includegraphics[width=\linewidth]{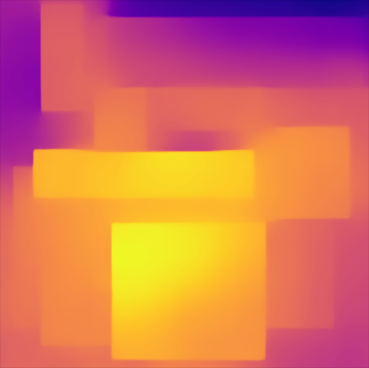}
    \end{subfigure}
    \begin{subfigure}[b]{0.38\linewidth}
        \centering
        \includegraphics[width=\linewidth]{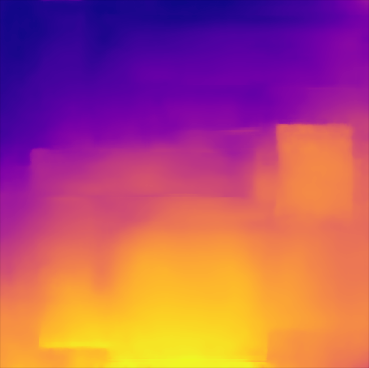}
    \end{subfigure}
    
    \caption{Results of state-of-the-art networks on our depth estimation problem, without retraining. First line: input and ground truth. Second line: result of MiDaS \cite{ranftl2019towards}, result of MergeNet \cite{miangoleh2021boosting}. The disappointing results of SOTA networks on a visually unambiguous image show that these networks are guided by hidden natural statistics, much more than by nonlocal geometric reasoning.
    }
    \label{fig:results_bignetworks}
\end{figure}

All of our results can be reproduced in less than one day on a single GPU. Both the code and dataset are available on \href{https://github.com/AdrienCourtois/neural-networks-properties}{GitHub}.

\section{Related work}

Multiple depth estimation datasets \cite{chen2020oasis,li2018megadepth,xian2018monocular,yu2020high} aim at training networks for real applications. 
The RDE dataset we introduce is a depth estimation task reduced to its simplest form, where only the strictly necessary cues are left for the network to understand the depth ordering of the scene.
Other synthetic datasets \cite{power2021grokking, tay2020long} have been proposed to analyze and quantify the effect of certain layers or training methods, allowing one to discover effects that would otherwise be impossible to unveil. Notably, synthetic datasets are commonly used for image quality evaluation \cite{kundu2018perceptual}. The Long-Range Arena \cite{tay2020long} was introduced to evaluate the long-range capabilities of Transformers \cite{vaswani2017attention}. 
While we share a similar objective, failure on such complex classification tasks cannot be easily linked to structural deficiencies. 
We aim at designing synthetic datasets for better understanding structures and not only assessing them.
The Color Code dataset is used to assess variants of Transformers, but both the RDE and the Centered Square dataset consist of images of small size but too large for the quadratic cost of Transformers. 
The RDE dataset shares similarities with the dead leaves model \cite{gousseau2003dead} and builds images composed of rectangles to create occlusion. 

In particular, the approach described in \cite{liu2018intriguing} is close to ours. The authors exhibit a property they want their network to have and design simple synthetic datasets to evaluate it. They find that their network does not have the property and propose a change in structure to solve the issue. In this work, we propose a generalization of this approach by providing a methodology to reproduce those steps for other properties.
In \cite{kokkinos2019pixel}, the authors also introduce a dataset to exhibit a property their layer has, but competing approaches do not. It can be argued however that the usage of noise introduces unneeded information and does not follow the Occam razor criterion. 

Different ways to improve the U-Net \cite{ronneberger2015u} have been proposed in the literature. For instance, $\text{U}^2\text{-Net}$ \cite{qin2020u2} provides a global receptive at each scale by including a U-Net at each scale of the U-Net. This method is state-of-the-art for figure-background segmentation. In \cite{alom2019recurrent} the receptive field and the amount of processing are increased by a recurrent network used at each scale of the encoder. We shall actually propose here a faster and lighter-weight approach by leveraging non-local layers to attain global receptive field at each scale. In \cite{zhu2020map} the authors  identify a receptive field issue with the U-Net. They propose to solve it by a novel structure processing all scales in parallel. The LambdaUNet \cite{ou2021lambdaunet} uses the Lambda layer \cite{bello2021lambdanetworks} in conjunction with the U-Net. It keeps the Lambda layer in its local formulation, while we shall change its receptive field to cover the entire image at once. Notably, the authors of \cite{wang2020non} introduced a network called ``Non-Local U-Net". They use a so-called ``non-local layer" \cite{wang2018non} similar to self-attention \cite{vaswani2017attention} to increase the receptive field of the U-Net. The resulting network is slow, as it is based on an operation with quadratic time and space complexity. In comparison, we shall explore a U-Net architecture that can be combined with a variety of non-local layers. The layers we choose to assess have a linear time and space complexity and can be trained and evaluated on a single GPU. 

A wide variety of non-local layers have been proposed in the literature. Many of them are based on self-attention or ``non-local networks" \cite{wang2018non}. Some layers aim at mimicking self-attention with a linear complexity \cite{wang2020linformer, shen2021efficient, choromanski2020rethinking, xiong2021nystr, kim2020fastformers, jaegle2021perceiver, huang2019ccnet, zhu2021long, xiong2021nystromformer}. We evaluated some of those layers but were not able to make them converge on our task, or they were exceedingly long to train. This suggests that they require heavy hyper-parameter tuning or the usage of multiple convergence tricks. In this work, we  explicitly chose to assess easy-to-use and easy-to-train layers. Other non-local layers \cite{bello2021lambdanetworks, gcnet, chen2019graph, dai2017deformable, hu2018gather, hu2018squeeze, tabernik2020spatially, zhu2019deformable} do not try to mimic self-attention. Any of them could be incorporated in our architecture. We shall  evaluate several of them.

Since its first introduction in \cite{vaswani2017attention}, different approaches have proposed different positional encoding methods. In \cite{chen2021demystifying}, it is pointed out that the positional encoding in its original formulation is not translation covariant. The authors propose to decorrelate the encoding of the absolute position with the encoding of the relative position. Their findings suggest that relative position alone is enough for some tasks in NLP. The original position encoding is a predefined sinusoidal function, and some works have focused on improving these functions \cite{liu2020learning, dehghani2018universal, lan2019albert}. Other approaches have been developed, see \cite{dufter2021position} for an overview. In this work, we use the Centered Square dataset to evaluate different positional encoding methods to be used inside the Lambda layer to ensure translation covariance.


\section{The methodology} \label{sec:dataset-algo-gt}

In this section, we describe the design requirements an unambiguous dataset must fulfill to assess whether or not a structure has a given property such as nonlocality, translation covariance, \textit{etc}. Such requirements  can only be fully enforced in a synthetic dataset, namely:

\smallskip\noindent\emph{Unambiguous ground-truth:} There must be no contradictory labels, no annotation problems nor cases where multiple labels are valid for the same input.

\smallskip\noindent\emph{Well-posedness:} The input contains enough information to solve the task.
There must exist a reconstruction algorithm able to deduce the exact ground truth from the input image. In other terms, reaching 100\% accuracy is theoretically possible. Note that, because of the inherent ambiguity of natural scenes, this property is not attainable with natural datasets.


\smallskip\noindent\emph{Focus on a specific network's property:} The network must be able to deduce the exact ground truth from the input image only if it has the assessed property (such as nonlocality, permutation invariance, etc.).  So the cues given to the network must be under full control, so that we know exactly which cues the network can use.
Note again that this property is not attainable with natural datasets, as they contain many  statistical cues that help compensate for a structural deficiency of the network.

In particular, we would like to stress out that an important requirement is \textit{simplicity}. The third property can only be enforced if the dataset is as simple as possible.

In the following, we describe the three datasets we used as an illustration of our methodology for, respectively, \textit{nonlocality}, \textit{translation covariance} and \textit{the ability to pass on information to every pixel of the same color}.

\subsection{The Rectangle Depth Estimation dataset}

\begin{figure}
    \centering
        \begin{subfigure}[b]{0.38\linewidth}
            \centering
            \includegraphics[width=\textwidth]{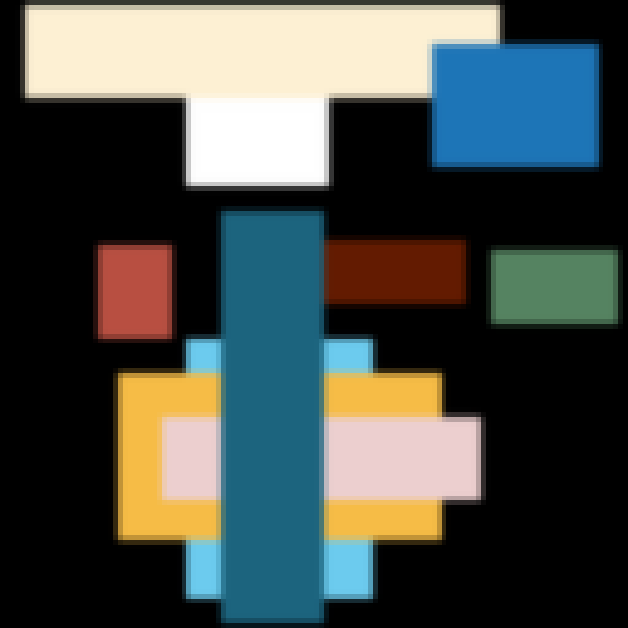}
        \end{subfigure}
        \begin{subfigure}[b]{0.38\linewidth}
            \centering
            \includegraphics[width=\textwidth]{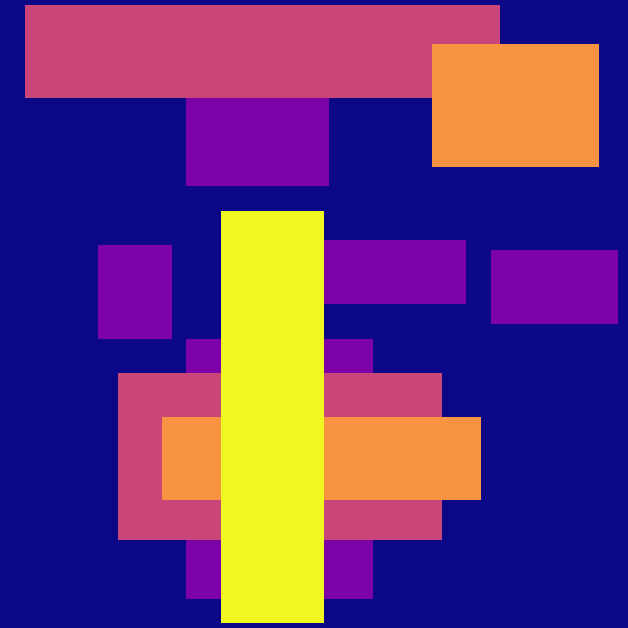}
        \end{subfigure}
        
        \begin{subfigure}[b]{0.38\linewidth}
            \centering
            \includegraphics[width=\textwidth]{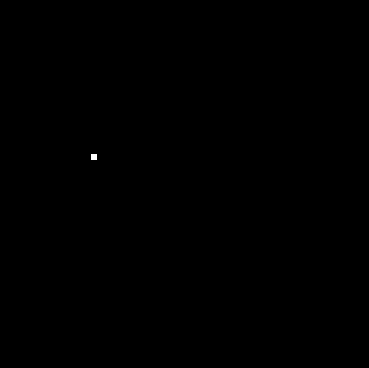}
        \end{subfigure}
        \begin{subfigure}[b]{0.38\linewidth}
            \centering
            \includegraphics[width=\textwidth]{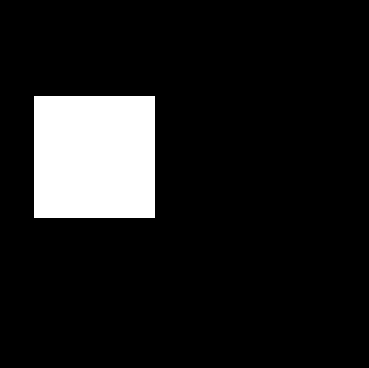}
        \end{subfigure}
    \caption{Top row: An example of image of the \DatasetName dataset and the associated ground truth. The brighter the color, the higher the number of rectangles that are beneath it. Bottom row: An example of image of the Centered Square dataset and the associated ground truth with $H=W=64$ and $w=21$.}
    \label{fig:example-dataset}
\end{figure}

The dataset consists in a depth estimation task where objects are replaced by simple rectangles. The rectangles can overlap and occlude one another, creating a spatial organization that naturally puts objects of top of others. To compute an unambiguous ground truth, our reconstruction algorithm  is based on three nonlocal cues: a) color similarity (all rectangles are monochromatic, thus can be recovered nonlocally); b) T-junctions, a local cue that propagates nonlocally; c) convexity, that leads to decide 
that a region occluded by another shape is to be inpainted as a convex shape and is therefore underneath the occluding shape. 
A full description of the algorithm is available in the supplementary materials. An example can be found in Figure \ref{fig:example-dataset}.

The task that needs to be solved is closely related to real-world depth estimation as it accurately reflects its main difficulties. When for example an object is partially occluded by others, it is divided into several components and the network must regroup the separated parts. This can only be done by recognizing the same color and/or detecting edge alignment.

To decide about the depth ordering, the network can only rely on T-junctions and convexity. These local cues need to be successfully detected and propagated at an arbitrary distance for understanding the geometrical organization of the scene. Indeed, the distance can be arbitrarily large as there is no upper limit on the size of the rectangles. Therefore, the network cannot overfit on local information, whereas in natural scenes it is easy for a network to differentiate, say, a tree from the background sky. Whilst these could be interesting priors, there is no guarantee that the network will not associate a depth value to each position or texture. Our dataset is designed so that it is not possible to associate a local patch to an absolute depth. All cues provide information about relative ordering between objects. The global depth can only emerge via a coherent global integration of these relative cues. In Figure \ref{fig:results_bignetworks} we show an example where state-of-the-art networks trained on natural images seem to heavily rely on local cues and natural statistics. We can for instance see that the bottom of the picture always seems to be brighter than the top, even though it makes no sense in this case. Of course, these networks being trained on natural data, it was to be expected that they would perform poorly on work  on our out of domain dataset. Nevertheless, this experiment is interesting because it shows statistical priors on the depth learnt. In our dataset, a failure cannot be attributed to a misunderstanding of the objects caused for instance by poor lightning conditions or noise. The dataset being fully unambiguous and its ground truth recoverable from geometric features in sight, failure can only be attributed to a poor geometrical understanding. This allows one to assess the ability for a network to compute non-local features (or, for the case of the U-Net, to efficiently use the multi-scale structure). This also suggests that any improvement on this dataset should be reflected on other depth estimation datasets.


\subsection{The Centered Square dataset} \label{sec:centeredsquare-dataset}
This dataset is designed to assess the translation covariance of a given positional encoding method. We use it to find the best method to use inside our Lambda layer. The input consists of an all-black $H \times W$ image where a single pixel is white. The associated ground truth is an all-black image with a white square of width $w$ centered around the white pixel. The training set consists of all the positions for the white pixel contained in the square of dimension $\frac{H}{2} \times \frac{W}{2}$ located in the center of the image. The test set is composed of all the other positions, except for the ones where the image's boundaries crop the ground truth square. This way, the network only learns the reconstruction property in the middle of the image and is evaluated on its ability to apply this property everywhere in the image. A network can only do it perfectly if it is translation equivariant. An example of input and label is shown in Figure \ref{fig:example-dataset}.

\subsection{The Color Code dataset} \label{sec:colorcode-dataset}
One of the limitations of the RDE dataset is that it  uses $10$ fixed colors for the entire dataset \textit{i.e.} for every image, the $10$ same colors are used to color the rectangles. We made this choice so the network could focus on the nonlocal reasoning, even if it implies overfitting on the fixed colors to overcome occlusion. In particular, we found that the baseline U-Net is not able to overcome occlusion even in this simplistic scenario. In our attempt to progressively bridge the gap between synthetic and real depth estimation, the next natural step is to change the colors for each image. In this scenario, a network must solve two tasks: first, it must use local and nonlocal cues to find a mapping between color and depth and secondly, it must pass on this depth to every pixel of this color.

As the second task is difficult in itself, we decided to study the performance of different layers on this task alone. This leads us to the introduction of the Color Code dataset. This third dataset aims to study the performance of a network which, given a mapping between colors and codes, must pass to every pixel the code corresponding to its color. More formally, for each input $k$ colors $c_1, \hdots, c_k$ are randomly sampled. For each color $c_i$, a code $z_i$ is randomly picked. Then, a mapping $\sigma: [\![1,k]\!] \to [\![1,N]\!]$ is sampled as well as a mask $m \in \{0,1\}^N$ such that the input is given by
$$x = \begin{pmatrix}c_{\sigma(1)} & \hdots & c_{\sigma(N)} \\ m_1 \cdot z_{\sigma(1)} & \hdots & m_N \cdot z_{\sigma(N)}\end{pmatrix},$$
and the associated ground truth is
$$y = \begin{pmatrix}z_{\sigma(1)} & \hdots & z_{\sigma(N)}\end{pmatrix}.$$
In other terms, for some positions the code is given and for others it is not. The goal of the network is to find where the code associated with a color has been given, retrieve it and propagate it to the right positions.

\section{Non-Local U-Net}
Our baseline for the depth estimation problem is a traditional U-Net \cite{ronneberger2015u} with concatenated skip connections. To limit the number of parameters, we kept the width at each scale constant and equal to 48 instead of doubling it after each down-sampling. In accordance with \cite{tan2021efficientnetv2}, we observed that this alleviated overfitting in multiple scenarios. The resulting network had 871 729 parameters. With the multiple skip connections and the hourglass structure, the U-Net is known to be stable to a variety of learning rates and training schedules. We modified this U-Net following the recipe of \cite{liu2022convnet}. This includes using GELU \cite{hendrycks2016gaussian}, LayerNorm \cite{layernorm} instead of BatchNorm \cite{batchnorm}, $7 \times 7$ grouped convolutions, the inverted bottleneck structure after each block \cite{sandler2018mobilenetv2} and LayerScale \cite{touvron2021going}.

To host nonlocal layers, we passed the input feature map at each scale into a local module and a nonlocal module. The two outputs were then concatenated, upsampled/downsampled and passed on to the next scale. The local module corresponds to the module of the original U-Net and the nonlocal module is the nonlocal layer. We refer the reader to Appendix A for further details.

All the networks we considered have five down-sampling operations. The smallest feature map has a $4 \times 4$ spatial extent. Therefore,  the receptive field of the baseline U-Net and all of its considered variants cover the entire image.

We tried four different non-local layers: the Lambda Layer \cite{bello2021lambdanetworks} (which is itself a variant of \cite{a2net}), the Global Context Layer \cite{gcnet}, Global Average Pooling, Deformable Convolutions \cite{zhu2019deformable}. We chose these over others because they could be applied at each scale and fit on a single GPU.

When it was not already the case, we embedded the nonlocal layer with a PreNorm \cite{nguyen2019transformers}, skip connections and an inverted bottleneck structure to process its output. We also fixed stability issues when discovered. We found  that these simple tweaks led to stabler convergence and better results. We modified the original Lambda layer to avoid using its positional encoding, which has a quadratic time/space complexity. More details can be found in Section \ref{sec:results-centeredsquare} and the supplementary.

\section{Experiments}

\subsection{Experiments on the \DatasetName dataset}

\subsubsection{Metrics}
We used three of the most commonly employed metrics for Monocular Depth Estimation tasks \cite{alhashim2018high,chen2016single,miangoleh2021boosting,ranftl2019towards}: the Root Mean Square Error (RMSE), the $\delta_{1.25}$ and the Ord metric. We also used the generalization gap as an indicator of how well the assessed networks generalize \cite{dai2021coatnet}. The RMSE was defined by
$$\text{RMSE}(\hat{y}, y) := \sqrt{\frac{1}{HW} \sum_{i,j} (\hat{y}_{i,j} - y_{i,j})^2},$$
where $\hat{y}$ is the prediction and $y$ the ground-truth. The percentage of pixels with $\delta_{1.25}$ is given by
$$\delta_{1.25} := \frac{1}{HW} \sum_{i,j} \mathbf{1}_{\{\max\left(\frac{\hat{y}_{i,j}}{y_{i,j}}, \frac{y_{i,j}}{\hat{y}_{i,j}}\right) > 1.25 \}}.$$
The ordinal loss consists in sampling 50,000 pairs of pixels $((i_1, j_1), (i_2, j_2))$ and for each of those pairs, compute:
$$l_i = \begin{cases}+1, & \text{ if } y_{i_1,j_1} / y_{i_2,j_2} \geq 1+\tau \\ -1, & \text{ if } y_{i_1,j_1} / y_{i_2,j_2} \leq \frac{1}{1+\tau} \\ 0, & \text{ otherwise.}\end{cases}$$
Using the same pairs, the equivalent quantity $\hat{l}$ is computed for the prediction. The ordinal loss is given by:
$$\text{Ord} := \frac{1}{|\mathcal{P}|} \sum_{i \in \mathcal{P}} \mathbf{1}_{\{ l_i \neq \hat{l}_i \}}.$$
Finally, we define the generalization gap as the difference between the value of the loss on the test set and on the train set at the end of the training.

In practice, we used $\tau = 0.03$ and all the networks were evaluated using the same set of pairs of pixels when computing the ordinal loss.

\subsubsection{Results}

\paragraph{Effect of ambiguity removal}
We trained the Non-Local U-Net with different nonlocal layers on the RDE dataset. This dataset was comprised of images of dimension $128 \times 128$. Most images featured 10 rectangles. The dataset was filtered so as to remove most  ambiguous cases, \textit{e.g.} when T-junctions are hidden by another square or when rectangle sides are aligned.  As an illustration of the need for an unambiguous dataset, we compare in Table \ref{tab:results-toy} the performance of the baseline network when trained on the unambiguous dataset and when trained on the same dataset but where we did not remove the ambiguous cases. The network trained on the unambiguous dataset is four times better than its counterpart.

\paragraph{Comparison}
We report in the upper part of Table \ref{tab:results-toy} the results of the different assessed nonlocal layers on the RDE dataset. They show the Lambda layer yielding the best performance for most metrics.

The deformable convolutions yielded the lowest performance. This is most likely due to the fact that it has the smallest width. Since this layer introduced a large number of parameters, we had to reduce the width so it had a number of parameters close to the baseline. Its width was $21$ when most layers had around $40$ channels per feature map.

Overall, even the simplest non-local layer yielded a noticeable improvement over the baseline U-Net. This supports the claim of \cite{qin2020u2} and \cite{zhu2020map} that the U-Net might be more local than expected. It seems that the more sophisticated the non-local layer, the better the results, which suggests that further improvement could come from still better nonlocal layers.

\begin{figure}
    \centering
    \begin{subfigure}[b]{0.38\linewidth}
        \centering
        \includegraphics[width=\linewidth]{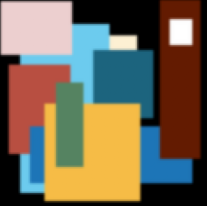}
    \end{subfigure}
    \begin{subfigure}[b]{0.38\linewidth}
        \centering
        \includegraphics[width=\linewidth]{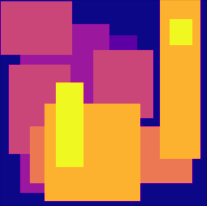}
    \end{subfigure}
    \begin{subfigure}[b]{0.38\linewidth}
        \centering
        \includegraphics[width=\linewidth]{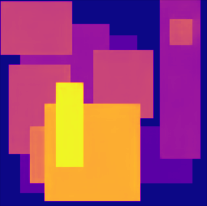}
    \end{subfigure}
    \begin{subfigure}[b]{0.38\linewidth}
        \centering
        \includegraphics[width=\linewidth]{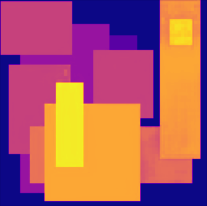}
    \end{subfigure}
    \caption{An example of case where the U-Net without nonlocal layers is not able to overcome occlusion. First line: input, ground truth; second line: output of the baseline U-Net, output of the Non-Local U-Net + Lambda + PE. To solve this case, the network must propagate the depth information it found on the left of the shape to the rest of the shape, with the help of the information of color.}
    \label{fig:example_propag_failure}
\end{figure}

Although the U-Net has a global receptive field, the way the information propagates inside it might be to blame. This information is fused locally, step by step, in the way of a diffusion process. This might explain why occlusions stop the  propagation of information from a piece of an occluded object to another, as can be observed in Figure \ref{fig:example_propag_failure}. See Section \ref{sec:visual-interp} for more details along with an illustration.

\begin{figure}
    \centering
    \begin{subfigure}[b]{0.38\linewidth}
        \centering
        \includegraphics[width=\linewidth]{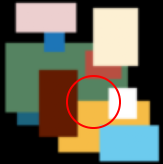}
    \end{subfigure}
    \begin{subfigure}[b]{0.38\linewidth}
        \centering
        \includegraphics[width=\linewidth]{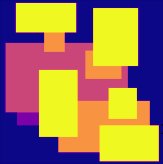}
    \end{subfigure}
    \begin{subfigure}[b]{0.38\linewidth}
        \centering
        \includegraphics[width=\linewidth]{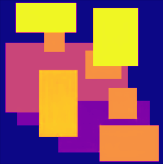}
    \end{subfigure}
    \begin{subfigure}[b]{0.38\linewidth}
        \centering
        \includegraphics[width=\linewidth]{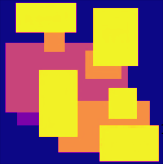}
    \end{subfigure}
    \caption{An example of case where the T-junctions between two rectangles are occluded. First line: input, ground truth; second line: output of the Non-Local U-Net + Lambda, output of the Non-Local U-Net + Lambda + PE. To solve this case, the network must compute the spatial extent of the occluded rectangles and determine which is on top. Incorporating a translation covariant positional encoding in the Lambda layer partially solved these problems.}
    \label{fig:example_occluded_tjunctions}
\end{figure}

When observing the cases where our best network failed, we observed that the network struggled in the case the T-junctions between two rectangles are occluded. In this case, the network needs to compute the spatial extent of each rectangle from the visible parts and understand that the extensions overlap. See Figure \ref{fig:example_occluded_tjunctions} for an illustration. Failure to handle such case suggests a problem with the positional encoding used within the Lambda layer, which leads us to the Centered Square dataset.

\begin{table*}
\begin{center}
\begin{tabular}{|l|c|c|c|c|c|c|}
\hline
Network & \# parameters & Test loss $\downarrow$ & Gen. gap $\downarrow$ & Ord $\downarrow$ & $\delta_{1.25}$ $\downarrow$ & RMSE $\downarrow$ \\
\hline\hline
Baseline* & 871 729 & 1.78 & - & 6.03 & 7.36 & 4.91 \\
Baseline & 871 729 & 0.72 & 0.36 & 1.79 & 1.98 & 1.95 \\
Global Context & 864 477 & 0.54 & 0.32 & 1.09 & 1.39 & 1.49 \\
Global Average Pooling & 883 049 & 0.60 & 0.31 & 1.47 & 1.51 & 1.71 \\
Deformable & 864 844 & 0.94 & 0.54 & 2.43 & 2.83 & 2.44 \\
Lambda & 871 945 & \underline{0.28} & \textbf{0.14} & \textbf{0.47} & \underline{0.64} & \underline{0.82} \\
\hline \hline
Lambda + PE & 871 945 & \textbf{0.25} & \underline{0.15} & \underline{0.50} & \textbf{0.58} & \textbf{0.77} \\
Lambda + PE + TT & 928 969 & 0.28 & 0.18 & 0.59 & 0.65 & 0.85 \\
\hline
\end{tabular}
\end{center}
\caption{Results on the \DatasetName dataset. All metrics are multiplied by 100 for readability. The best reported results are in bold and  the second best is underlined. Note that the more sophisticated the non-local layer, the better the results.
The model marked with an asterisk (*) was trained on the ambiguous version of the training dataset and evaluated on the unambiguous one. In particular, the reported test loss is the one computed on the unambiguous dataset.
}
\label{tab:results-toy}
\end{table*}

\subsection{Results on the Centered Square dataset}\label{sec:results-centeredsquare}
For this set of experiments, we used the Centered Square dataset presented in Section \ref{sec:centeredsquare-dataset} with $H=W=64$ and a square width of $w=21$. The dataset was composed of 484 training images and 1,452 test images. We evaluated our network by computing the IOU over the test set. Further training details are given in the supplementary.

As the goal of this dataset was to evaluate the positional encoding method, the network to be trained was reduced to its simplest form. Indeed, using a multiscale structure could bias the interpretation of the results. On this task, we trained  a network made of one $1 \times 1$ convolution, followed by the Lambda layer using the positional encoding method being investigated and by another $1 \times 1$ convolution. 

In NLP, the Transformer-based architecture almost exclusively relies on positional encoding strategies to encode the relative and absolute positions of words in a sentence. The original Lambda layer is inspired by the Transformer architecture but the original positional encoding of the lambda layer can be very costly in terms of both parameters and computations. Therefore, we decided to replace it with the cosine positional embedding presented in \cite{vaswani2017attention}. Our first approach was to simply add the positional encoding to the input feature map and pass the result through the Lambda layer. As pointed out in \cite{chen2021demystifying}, this leaks the absolute position of the pixel, which could be detrimental for our task. We therefore decided to decorrelate the positional encoding from the rest of the layer by adapting the method of \cite{chen2021demystifying} to the lambda layer. Formally, given a predefined positional encoding $P \in \mathbb{R}^{C \times N}$, an input feature map $x \in \mathbb{R}^{C_{\text{in}} \times N}$ and learnable matrices $K \in \mathbb{R}^{M \times C_{\text{in}}}$, $V \in \mathbb{R}^{C_{\text{out}} \times C_{\text{in}}}$, $Q \in \mathbb{R}^{M \times C_{\text{in}}}$, $A \in \mathbb{R}^{C_{\text{out}} \times M}$ we compute
$$\bar{K} = \textsc{softmax}_{N} (K x) \in \mathbb{R}^{M \times N},$$
$$\lambda_{\text{content}} = \bar{K} (Vx)^T \in \mathbb{R}^{M \times C_{\text{out}}},$$
$$\lambda_{\text{pos}} = A \bar{K} P^T \in \mathbb{R}^{C_{\text{out}} \times C},$$
$$y_{\text{pos}} = \lambda_{\text{pos}} P \in \mathbb{R}^{C_{\text{out}} \times N},$$
$$y_{\text{content}} = \lambda_{\text{content}}^T Qx \in \mathbb{R}^{C_{\text{out}} \times N},$$
and the output of the Lambda layer is given by $y = y_{\text{content}} + y_{\text{pos}}$. We refer to this method as ``Decor.". Finally, we noted that the cosine positional encoding of \cite{vaswani2017attention} is of the form
$$P_{c,n} = \begin{cases}\cos(w_{k} n) & \text{ if } c = 2k \\ \sin(w_{k} n) & \text{ if } c = 2k+1\end{cases},$$
and we investigated if another choice of the sequence $(w_c)_{c \in [\![1,C/2]\!]}$ could yield better results. This led us to introduce Fourier coefficients $w_c = 2\pi c / (2C)$, $c=1,\hdots,C/2$. We refer to this method as ``Fourier".

In Table \ref{tab:results-centeredsquare}, we compare these methods with the use of CoordConv \cite{liu2018intriguing} instead of  regular convolutions within the Lambda layer, and with the translation covariant version of the positional encoding proposed in the original Lambda layer with different widths $R$. Notably, we found that the only truly translation covariant approaches were the one that used the ``Decor." mechanism. In particular, using the Fourier positional encoding alongside the ``Decor." mechanism yields a perfect score on the dataset. We tested our variant of the Lambda layer with this new positional encoding method (Lambda + PE) on the RDE dataset. This modification moderately improved on the final performance as reported in Table \ref{tab:results-toy}.

\begin{table*}
\begin{center}
\begin{tabular}{|l|c|c|c|}
\hline
Method & \# parameters & Train IOU & Test IOU \\
\hline \hline 
Cosine + Sum + QKV & 182 018 & \underline{98.70\%} & 24.71\% \\
Cosine + Sum + QV  & 182 018 & 98.67\% & 35.66\% \\
Cosine + Decor.    & 182 018 & 84.78\% & \underline{84.79\%} \\
Fourier + Decor.   & 182 018 & \textbf{100.0\%} & \textbf{100.0\%} \\
CoordConv          & 184 066 & 80.71\% & 56.16\% \\
Lambda ($R=7$)     & 182 818 & 11.11\% & 11.11\% \\
Lambda ($R=19$)    & 187 810 & 19.50\% & 14.81\% \\
\hline
\end{tabular}
\end{center}
\caption{Results on the Centered Square dataset with input dimension $H=W=64$ and square width $w=21$. The best reported results are in bold and the second best are underlined. The ``Decor." method is the most consistent as it performs almost identically on the train and test sets.}
\label{tab:results-centeredsquare}
\end{table*}

\subsection{Results on the ColorCode dataset}
For this set of experiments, we used the Color Code dataset presented in Section \ref{sec:colorcode-dataset}. The input dimension  was $N=128$, the number of different colors per input  $k=10$ and the proportion of masked inputs $50\%$. We trained our networks with 20,000 training images and evaluated them on a separate test set comprised of 10,000 testing images. The results are presented in Table \ref{tab:results-colorcode}.

The spatial dimension of the input being low, we chose to evaluate the smallest versions of the linear-cost approximation of the self-attention. As we only needed to investigate layer capabilities, we  reduced the trained networks to their simplest form: a $1 \times 1$ convolution, followed by three instances of the assessed layer, followed by another $1 \times 1$ convolution. See the supplementary for more details.

The used metric was the mean accuracy of the masked codes across the training dataset \textit{i.e.}
$$\frac{\sum_{i=1}^N (1 - m_i) \cdot \mathbf{1}_{\{z_{\sigma(i)} = \hat{z}_i\}}}{\sum_{i=1}^N (1 - m_i)},$$
where $\hat{z}$ is the network's prediction.

Notably, all self-attention variants perform on par. The overall performance of the Transformer suggests that some ambiguity was left in the dataset. This ambiguity was probably due to some colors not being easily distinguishable. Humans would also fail in some cases as it is hard to tell apart the $256^3$ different colors. The failure of the MLP-Mixer on this task seems to indicate that multilayer perceptrons are not always a good replacement for self-attention, even if they are on image classification.

The only attention map computed in the Lambda layer depends on the comparison between the input feature map and a learned matrix. Inspired by the mechanism of self-attention, we improved this attention map by switching the learned matrix for a matrix computed based on the input feature map. This amounts to iterating the Lambda layer \textbf{twice}, yielding our variant named \textit{Lambda + TT}. We refer the reader to the supplementary for further details. On the Color Code dataset, this slight modification of the Lambda layer yields the second best performance. 
We tested this layer onto the RDE dataset and surprisingly, we found a decrease in performance as reported in Table \ref{tab:results-toy}. The lack of improvement could be due to this additional mechanism being not needed since there are only 10 colors in the entire dataset. It could also be due to the fact it was trained for 50 epochs while it could have benefited from a longer training.

\begin{table}
\begin{center}
\begin{tabular}{|l|c|c|c|}
\hline
Network & \# parameters & Test metric \\
\hline \hline 
Transformer      & 2 373 379 & \textbf{99.87\%} \\
MLP-Mixer        & 1 979 011 & 69.77\% \\
Nyströmformer-32 & 2 176 003 & 99.66\% \\
Linformer-32     & 2 208 771 & 99.15\% \\
Reformer-32      & 2 176 003 & 77.59\% \\
Lambda           & 2 175 235 & 99.52\% \\
Lambda + TT      & 2 569 987 & \underline{99.85\%} \\
\hline
\end{tabular}
\end{center}
\caption{Results on the Color Code dataset with $N=128$ positions and $k=10$ different colors for each input. In particular, the Transformer seems to have reached the maximum possible metric on this dataset.}
\label{tab:results-colorcode}
\end{table}

\section{Limitations, conclusion and future works}
\label{sec:visual-interp}

\subsection{Limitations}
This work is about better understanding the properties of neural structures. The goal of the methodology is to pinpoint the properties of each given structure. To this end, we remove all the complex unknown statistical cues inherent to natural images. The final goal is that, given a task to solve, a practitioner will first identify the properties needed to solve the task and will choose the components of the network accordingly. This nonetheless raises several issues. First, there is no guarantee that simple properties are easily identifiable for every task. Secondly, there is no guarantee that if we mix multiple structures with different properties, the resulting structure will have the properties of its components. Thirdly, even if we had managed to find a structure with all of the desired properties, it might be that it doesn't transfer to natural images.

Furthermore, all of this work is constrained by the optimization process: it might be that a structure that does not work on a given dataset would yield a very good result if trained using a different training recipe. 

\subsection{Conclusion and future work}
We attempted to design  synthetic datasets as tools  to compare and improve neural networks. Very controlled datasets like RDE might play the role that was formerly given in signal processing to the impulses that were fed to a black box to obtain an impulse response. Here, the goal is to keep interpretable results that link network structure changes to performance gains. We claim that such interpretations can hardly be obtained with natural annotated datasets.

We plan to expend RDE to more  general scenes while keeping its statistical neutrality. The current dataset does not address the non-local problem of detecting the main colors (the ten colors were fixed once and for all in the dataset). We plan to vary the number of rectangles, then to authorize more varied shapes, finally to endow them with textures, so as to keep the synthetic dataset visually interpretable, statistically neutral, but ever closer in complexity to a natural scene.  

{\small
\bibliographystyle{ieee_fullname}
\bibliography{egbib}

\begin{thebibliography}{10}\itemsep=-1pt

\bibitem{alhashim2018high}
Ibraheem Alhashim and Peter Wonka.
\newblock High quality monocular depth estimation via transfer learning.
\newblock {\em arXiv preprint arXiv:1812.11941}, 2018.

\bibitem{alom2019recurrent}
Md~Zahangir Alom, Chris Yakopcic, Mahmudul Hasan, Tarek~M Taha, and Vijayan~K
  Asari.
\newblock Recurrent residual u-net for medical image segmentation.
\newblock {\em Journal of Medical Imaging}, 6(1):014006, 2019.

\bibitem{layernorm}
Jimmy~Lei Ba, Jamie~Ryan Kiros, and Hinton Geoffrey~E.
\newblock Layer normalization.
\newblock {\em arXiv preprint arXiv:1607.06450}, 2016.

\bibitem{bello2021lambdanetworks}
Irwan Bello.
\newblock Lambdanetworks: Modeling long-range interactions without attention.
\newblock {\em arXiv preprint arXiv:2102.08602}, 2021.

\bibitem{gcnet}
Yue Cao, Jiarui Xu, Stephen Lin, Fangyun Wei, and Han Hu.
\newblock Global context networks.
\newblock In {\em IEEE Transactions on Pattern Analysis and Machine
  Intelligence}, 2020.

\bibitem{chen2021demystifying}
Pu-Chin Chen, Henry Tsai, Srinadh Bhojanapalli, Hyung~Won Chung, Yin-Wen Chang,
  and Chun-Sung Ferng.
\newblock Demystifying the better performance of position encoding variants for
  transformer.
\newblock {\em arXiv preprint arXiv:2104.08698}, 2021.

\bibitem{chen2016single}
Weifeng Chen, Zhao Fu, Dawei Yang, and Jia Deng.
\newblock Single-image depth perception in the wild.
\newblock {\em Advances in neural information processing systems}, 29:730--738,
  2016.

\bibitem{chen2020oasis}
Weifeng Chen, Shengyi Qian, David Fan, Noriyuki Kojima, Max Hamilton, and Jia
  Deng.
\newblock Oasis: A large-scale dataset for single image 3d in the wild.
\newblock In {\em Proceedings of the IEEE/CVF Conference on Computer Vision and
  Pattern Recognition}, pages 679--688, 2020.

\bibitem{a2net}
Yunpeng Chen, Yannis Kalantidis, Jianshu Li, Shuicheng Yan, and Jiashi Feng.
\newblock A2-nets: Double attention networks.
\newblock {\em arXiv preprint arXiv:1810.11579}, 2018.

\bibitem{chen2019graph}
Yunpeng Chen, Marcus Rohrbach, Zhicheng Yan, Yan Shuicheng, Jiashi Feng, and
  Yannis Kalantidis.
\newblock Graph-based global reasoning networks.
\newblock In {\em Proceedings of the IEEE/CVF Conference on Computer Vision and
  Pattern Recognition}, pages 433--442, 2019.

\bibitem{choromanski2020rethinking}
Krzysztof Choromanski, Valerii Likhosherstov, David Dohan, Xingyou Song,
  Andreea Gane, Tamas Sarlos, Peter Hawkins, Jared Davis, Afroz Mohiuddin,
  Lukasz Kaiser, et~al.
\newblock Rethinking attention with performers.
\newblock {\em arXiv preprint arXiv:2009.14794}, 2020.

\bibitem{dai2017deformable}
Jifeng Dai, Haozhi Qi, Yuwen Xiong, Yi Li, Guodong Zhang, Han Hu, and Yichen
  Wei.
\newblock Deformable convolutional networks.
\newblock In {\em Proceedings of the IEEE international conference on computer
  vision}, pages 764--773, 2017.

\bibitem{dai2021coatnet}
Zihang Dai, Hanxiao Liu, Quoc~V Le, and Mingxing Tan.
\newblock Coatnet: Marrying convolution and attention for all data sizes.
\newblock {\em arXiv preprint arXiv:2106.04803}, 2021.

\bibitem{dehghani2018universal}
Mostafa Dehghani, Stephan Gouws, Oriol Vinyals, Jakob Uszkoreit, and {\L}ukasz
  Kaiser.
\newblock Universal transformers.
\newblock {\em arXiv preprint arXiv:1807.03819}, 2018.

\bibitem{dufter2021position}
Philipp Dufter, Martin Schmitt, and Hinrich Sch{\"u}tze.
\newblock Position information in transformers: An overview.
\newblock {\em arXiv preprint arXiv:2102.11090}, 2021.

\bibitem{gousseau2003dead}
Yann Gousseau and Fran{\c{c}}ois Roueff.
\newblock The dead leaves model: general results and limits at small scales.
\newblock {\em arXiv preprint math/0312035}, 2003.

\bibitem{hendrycks2016gaussian}
Dan Hendrycks and Kevin Gimpel.
\newblock Gaussian error linear units (gelus).
\newblock {\em arXiv preprint arXiv:1606.08415}, 2016.

\bibitem{hu2018gather}
Jie Hu, Li Shen, Samuel Albanie, Gang Sun, and Andrea Vedaldi.
\newblock Gather-excite: Exploiting feature context in convolutional neural
  networks.
\newblock {\em arXiv preprint arXiv:1810.12348}, 2018.

\bibitem{hu2018squeeze}
Jie Hu, Li Shen, and Gang Sun.
\newblock Squeeze-and-excitation networks.
\newblock In {\em Proceedings of the IEEE conference on computer vision and
  pattern recognition}, pages 7132--7141, 2018.

\bibitem{huang2019ccnet}
Zilong Huang, Xinggang Wang, Lichao Huang, Chang Huang, Yunchao Wei, and Wenyu
  Liu.
\newblock Ccnet: Criss-cross attention for semantic segmentation.
\newblock In {\em Proceedings of the IEEE/CVF International Conference on
  Computer Vision}, pages 603--612, 2019.

\bibitem{batchnorm}
Sergey Ioffe and Christian Szegedy.
\newblock Batch normalization: Accelerating deep network training by reducing
  internal covariate shift.
\newblock {\em International conference on machine learning}, pages 448--456,
  2015.

\bibitem{jaegle2021perceiver}
Andrew Jaegle, Sebastian Borgeaud, Jean-Baptiste Alayrac, Carl Doersch, Catalin
  Ionescu, David Ding, Skanda Koppula, Daniel Zoran, Andrew Brock, Evan
  Shelhamer, et~al.
\newblock Perceiver io: A general architecture for structured inputs \&
  outputs.
\newblock {\em arXiv preprint arXiv:2107.14795}, 2021.

\bibitem{jumper2021highly}
John Jumper, Richard Evans, Alexander Pritzel, Tim Green, Michael Figurnov,
  Olaf Ronneberger, Kathryn Tunyasuvunakool, Russ Bates, Augustin
  {\v{Z}}{\'\i}dek, Anna Potapenko, et~al.
\newblock Highly accurate protein structure prediction with alphafold.
\newblock {\em Nature}, 596(7873):583--589, 2021.

\bibitem{kim2020fastformers}
Young~Jin Kim and Hany~Hassan Awadalla.
\newblock Fastformers: Highly efficient transformer models for natural language
  understanding.
\newblock {\em arXiv preprint arXiv:2010.13382}, 2020.

\bibitem{kokkinos2019pixel}
Filippos Kokkinos, Ioannis Marras, Matteo Maggioni, Gregory Slabaugh, and
  Stefanos Zafeiriou.
\newblock Pixel adaptive filtering units.
\newblock {\em arXiv preprint arXiv:1911.10581}, 2019.

\bibitem{kundu2018perceptual}
Debarati Kundu, Lark~Kwon Choi, Alan~C Bovik, and Brian~L Evans.
\newblock Perceptual quality evaluation of synthetic pictures distorted by
  compression and transmission.
\newblock {\em Signal Processing: Image Communication}, 61:54--72, 2018.

\bibitem{lan2019albert}
Zhenzhong Lan, Mingda Chen, Sebastian Goodman, Kevin Gimpel, Piyush Sharma, and
  Radu Soricut.
\newblock Albert: A lite bert for self-supervised learning of language
  representations.
\newblock {\em arXiv preprint arXiv:1909.11942}, 2019.

\bibitem{li2018megadepth}
Zhengqi Li and Noah Snavely.
\newblock Megadepth: Learning single-view depth prediction from internet
  photos.
\newblock In {\em Proceedings of the IEEE Conference on Computer Vision and
  Pattern Recognition}, pages 2041--2050, 2018.

\bibitem{liu2018intriguing}
Rosanne Liu, Joel Lehman, Piero Molino, Felipe Petroski~Such, Eric Frank, Alex
  Sergeev, and Jason Yosinski.
\newblock An intriguing failing of convolutional neural networks and the
  coordconv solution.
\newblock {\em Advances in neural information processing systems}, 31, 2018.

\bibitem{liu2020learning}
Xuanqing Liu, Hsiang-Fu Yu, Inderjit Dhillon, and Cho-Jui Hsieh.
\newblock Learning to encode position for transformer with continuous dynamical
  model.
\newblock In {\em International Conference on Machine Learning}, pages
  6327--6335. PMLR, 2020.

\bibitem{liu2022convnet}
Zhuang Liu, Hanzi Mao, Chao-Yuan Wu, Christoph Feichtenhofer, Trevor Darrell,
  and Saining Xie.
\newblock A convnet for the 2020s.
\newblock {\em arXiv preprint arXiv:2201.03545}, 2022.

\bibitem{miangoleh2021boosting}
S~Mahdi~H Miangoleh, Sebastian Dille, Long Mai, Sylvain Paris, and Yagiz Aksoy.
\newblock Boosting monocular depth estimation models to high-resolution via
  content-adaptive multi-resolution merging.
\newblock In {\em Proceedings of the IEEE/CVF Conference on Computer Vision and
  Pattern Recognition}, pages 9685--9694, 2021.

\bibitem{nguyen2019transformers}
Toan~Q Nguyen and Julian Salazar.
\newblock Transformers without tears: Improving the normalization of
  self-attention.
\newblock {\em arXiv preprint arXiv:1910.05895}, 2019.

\bibitem{ou2021lambdaunet}
Yanglan Ou, Ye Yuan, Xiaolei Huang, Kelvin Wong, John Volpi, James~Z Wang, and
  Stephen~TC Wong.
\newblock Lambdaunet: 2.5 d stroke lesion segmentation of diffusion-weighted mr
  images.
\newblock {\em arXiv preprint arXiv:2104.13917}, 2021.

\bibitem{paul2015review}
Sandeep Paul, Lotika Singh, et~al.
\newblock A review on advances in deep learning.
\newblock In {\em 2015 IEEE Workshop on Computational Intelligence: Theories,
  Applications and Future Directions (WCI)}, pages 1--6. IEEE, 2015.

\bibitem{power2021grokking}
Alethea Power, Yuri Burda, Harri Edwards, Igor Babuschkin, and Vedant Misra.
\newblock Grokking: Generalization beyond overfitting on small algorithmic
  datasets.
\newblock In {\em ICLR MATH-AI Workshop}, 2021.

\bibitem{qin2020u2}
Xuebin Qin, Zichen Zhang, Chenyang Huang, Masood Dehghan, Osmar~R Zaiane, and
  Martin Jagersand.
\newblock U2-net: Going deeper with nested u-structure for salient object
  detection.
\newblock {\em Pattern Recognition}, 106:107404, 2020.

\bibitem{ranftl2019towards}
Ren{\'e} Ranftl, Katrin Lasinger, David Hafner, Konrad Schindler, and Vladlen
  Koltun.
\newblock Towards robust monocular depth estimation: Mixing datasets for
  zero-shot cross-dataset transfer.
\newblock {\em arXiv preprint arXiv:1907.01341}, 2019.

\bibitem{ronneberger2015u}
Olaf Ronneberger, Philipp Fischer, and Thomas Brox.
\newblock U-net: Convolutional networks for biomedical image segmentation.
\newblock In {\em International Conference on Medical image computing and
  computer-assisted intervention}, pages 234--241. Springer, 2015.

\bibitem{sandler2018mobilenetv2}
Mark Sandler, Andrew Howard, Menglong Zhu, Andrey Zhmoginov, and Liang-Chieh
  Chen.
\newblock Mobilenetv2: Inverted residuals and linear bottlenecks.
\newblock In {\em Proceedings of the IEEE conference on computer vision and
  pattern recognition}, pages 4510--4520, 2018.

\bibitem{shen2021efficient}
Zhuoran Shen, Mingyuan Zhang, Haiyu Zhao, Shuai Yi, and Hongsheng Li.
\newblock Efficient attention: Attention with linear complexities.
\newblock In {\em Proceedings of the IEEE/CVF Winter Conference on Applications
  of Computer Vision}, pages 3531--3539, 2021.

\bibitem{tabernik2020spatially}
Domen Tabernik, Matej Kristan, and Ale{\v{s}} Leonardis.
\newblock Spatially-adaptive filter units for compact and efficient deep neural
  networks.
\newblock {\em International Journal of Computer Vision}, 128(8):2049--2067,
  2020.

\bibitem{tan2021efficientnetv2}
Mingxing Tan and Quoc~V Le.
\newblock Efficientnetv2: Smaller models and faster training.
\newblock {\em arXiv preprint arXiv:2104.00298}, 2021.

\bibitem{tay2020long}
Yi Tay, Mostafa Dehghani, Samira Abnar, Yikang Shen, Dara Bahri, Philip Pham,
  Jinfeng Rao, Liu Yang, Sebastian Ruder, and Donald Metzler.
\newblock Long range arena: A benchmark for efficient transformers.
\newblock {\em arXiv preprint arXiv:2011.04006}, 2020.

\bibitem{touvron2021going}
Hugo Touvron, Matthieu Cord, Alexandre Sablayrolles, Gabriel Synnaeve, and
  Herv{\'e} J{\'e}gou.
\newblock Going deeper with image transformers.
\newblock In {\em Proceedings of the IEEE/CVF International Conference on
  Computer Vision}, pages 32--42, 2021.

\bibitem{vaswani2017attention}
Ashish Vaswani, Noam Shazeer, Niki Parmar, Jakob Uszkoreit, Llion Jones,
  Aidan~N Gomez, {\L}ukasz Kaiser, and Illia Polosukhin.
\newblock Attention is all you need.
\newblock In {\em Advances in neural information processing systems}, pages
  5998--6008, 2017.

\bibitem{wang2020linformer}
Sinong Wang, Belinda~Z Li, Madian Khabsa, Han Fang, and Hao Ma.
\newblock Linformer: Self-attention with linear complexity.
\newblock {\em arXiv preprint arXiv:2006.04768}, 2020.

\bibitem{wang2018non}
Xiaolong Wang, Ross Girshick, Abhinav Gupta, and Kaiming He.
\newblock Non-local neural networks.
\newblock In {\em Proceedings of the IEEE conference on computer vision and
  pattern recognition}, pages 7794--7803, 2018.

\bibitem{wang2020non}
Zhengyang Wang, Na Zou, Dinggang Shen, and Shuiwang Ji.
\newblock Non-local u-nets for biomedical image segmentation.
\newblock In {\em Proceedings of the AAAI Conference on Artificial
  Intelligence}, volume~34, pages 6315--6322, 2020.

\bibitem{xian2018monocular}
Ke Xian, Chunhua Shen, Zhiguo Cao, Hao Lu, Yang Xiao, Ruibo Li, and Zhenbo Luo.
\newblock Monocular relative depth perception with web stereo data supervision.
\newblock In {\em Proceedings of the IEEE Conference on Computer Vision and
  Pattern Recognition}, pages 311--320, 2018.

\bibitem{xiong2021nystromformer}
Yunyang Xiong, Zhanpeng Zeng, Rudrasis Chakraborty, Mingxing Tan, Glenn Fung,
  Yin Li, and Vikas Singh.
\newblock Nystr{\"o}mformer: A nyst{\"o}m-based algorithm for approximating
  self-attention.
\newblock In {\em Proceedings of the... AAAI Conference on Artificial
  Intelligence. AAAI Conference on Artificial Intelligence}, volume~35, page
  14138. NIH Public Access, 2021.

\bibitem{xiong2021nystr}
Yunyang Xiong, Zhanpeng Zeng, Rudrasis Chakraborty, Mingxing Tan, Glenn Fung,
  Yin Li, and Vikas Singh.
\newblock Nyströmformer: A nyström-based algorithm for approximating
  self-attention.
\newblock {\em arXiv preprint arXiv:2102.03902}, 2021.

\bibitem{yu2020high}
Haichao Yu, Ning Xu, Zilong Huang, Yuqian Zhou, and Humphrey Shi.
\newblock High-resolution deep image matting.
\newblock {\em arXiv preprint arXiv:2009.06613}, 2020.

\bibitem{zhu2021long}
Chen Zhu, Wei Ping, Chaowei Xiao, Mohammad Shoeybi, Tom Goldstein, Anima
  Anandkumar, and Bryan Catanzaro.
\newblock Long-short transformer: Efficient transformers for language and
  vision.
\newblock {\em arXiv preprint arXiv:2107.02192}, 2021.

\bibitem{zhu2020map}
Qing Zhu, Cheng Liao, Han Hu, Xiaoming Mei, and Haifeng Li.
\newblock Map-net: Multiple attending path neural network for building
  footprint extraction from remote sensed imagery.
\newblock {\em IEEE Transactions on Geoscience and Remote Sensing}, 2020.

\bibitem{zhu2019deformable}
Xizhou Zhu, Han Hu, Stephen Lin, and Jifeng Dai.
\newblock Deformable convnets v2: More deformable, better results.
\newblock In {\em Proceedings of the IEEE/CVF Conference on Computer Vision and
  Pattern Recognition}, pages 9308--9316, 2019.

\end{thebibliography}
}

\end{document}